\documentclass[a4paper,10pt,onecolumn]{article}
\usepackage[textwidth=14cm]{geometry}
\usepackage[utf8]{inputenc}
\usepackage{amsmath}
\usepackage{amsfonts}
\usepackage{amssymb}
\usepackage{cite}
\usepackage{graphicx}
\usepackage{microtype}
\usepackage{enumitem}
\usepackage{algorithm}
\usepackage{xcolor}
\usepackage[noend]{algpseudocode}
\usepackage{bm}
\usepackage{color}
\usepackage{multirow}
\usepackage{bbold}
\usepackage[font=small, labelfont = sc]{caption}
\usepackage{authblk}
\usepackage[hidelinks]{hyperref}

\algnewcommand\algorithmicinput{\textbf{Initialization:}}
\algnewcommand\init{\item[\algorithmicinput]}
\algnewcommand\algorithmicawake{\textsf{\textit{{AWAKE}}}}
\algnewcommand\awake{\item[\algorithmicawake]}
\algnewcommand\algorithmicidle{\textsf{\textit{{IDLE}}}}
\algnewcommand\idle{\item[\algorithmicidle]}
\newcommand{\broad}{{\small \textbf{BROADCAST }}}

 \newcommand{\DD}{\mathcal{D}}
\newcommand{\EE}{\mathcal{E}}
\newcommand{\GG}{\mathcal{G}} 
 
\newcommand{\NN}{\mathcal{N}}

\newcommand{\VV}{\mathcal{V}}

\newcommand{\st}{\text{subject to }}

\newcommand{\m}{\mathop{\rm minimize}}

\newcommand{\p}{{p}}

\newcommand{\LL}{{\theta}}

\newcommand{\rhoij}{\rho_{ij}}

\newcommand{\peqi}{\varrho_{i}}
\newcommand{\pini}{\zeta_{i}}

\newcommand{\tLag}{\tilde{\mathcal{L}}}
\newcommand{\Lag}{\mathcal{L}}

\newcommand{\wis}{w^s{\mid}_i}

\newcommand{\wjs}{w^s{\mid}_j}

\newcommand{\w}{{w}}
\newcommand{\si}{s{\mid}_i}

\makeatletter
\newcommand{\pushright}[1]{\ifmeasuring@#1\else\omit\hfill$\displaystyle#1$\fi\ignorespaces}
\newcommand{\pushleft}[1]{\ifmeasuring@#1\else\omit$\displaystyle#1$\hfill\fi\ignorespaces}
\makeatother

\makeatletter
\def\old@comma{,}
\catcode`\,=13
\def,{%
  \ifmmode%
    \old@comma\discretionary{}{}{}%
  \else%
    \old@comma%
  \fi%
}
\makeatother

\graphicspath{{figs/},{.}}

\begin{document}
\title{Asynchronous Distributed Learning from Constraints\footnote{
\textcopyright 2019 IEEE.  Personal use of this material is permitted.  Permission from IEEE must be obtained for all other uses, in any current or future media, including reprinting/republishing this material for advertising or promotional purposes, creating new collective works, for resale or redistribution to servers or lists, or reuse of any copyrighted component of this work in other works.
}}
\author[]{Francesco~Farina}
\author[]{Stefano~Melacci}
\author[]{Andrea~Garulli}
\author[]{Antonio~Giannitrapani}
\affil[]{Dipartimento di Ingegneria dell'Informazione e Scienze Matematiche, Universit{\`a} di Siena, Siena, Italy.}
\date{}

\maketitle
\begin{abstract}
In this paper, the extension of the framework of Learning from Constraints (LfC) to a distributed setting where multiple parties, connected over the network, contribute to the learning process is studied. LfC relies on the generic notion of ``constraint'' to inject knowledge into the learning problem and, due to its generality, it deals with possibly nonconvex constraints, enforced either in a hard or soft way. Motivated by recent progresses in the field of distributed and constrained nonconvex optimization, we apply the (distributed) Asynchronous Method of Multipliers (ASYMM) to LfC. The study shows that such a method allows us to support scenarios where selected constraints (i.e., knowledge), data, and outcomes of the learning process can be locally stored in each computational node without being shared with the rest of the network, opening the road to further investigations into privacy-preserving LfC. Constraints act as a bridge between what is shared over the net and what is private to each node and no central authority is required. We demonstrate the applicability of these ideas in two distributed real-world settings in the context of digit recognition and document classification.
\end{abstract}
\section{Introduction}
\label{sec:Intro}
The generic framework of Learning from Constraints (LfC) \cite{gori2017machine,gnecco2015foundations,gnecco2015learning} reframes the learning process in a context that is described by a collection of constraints. Such constraints are the mean that is used to inject knowledge into the learning process and they represent different aspects of the task at hand. 
The goal of LfC is to learn a vector function $f$ (classifier, regressor, etc.) by solving a constrained optimization problem where $f$ is required to maximize some regularity conditions in the space to which it belongs \cite{gnecco2015foundations}.
Different types of knowledge can be exploited in LfC, including the ones that are represented using First-Order Logic (FOL) formulas \cite{gori2013constraint}. For example, knowledge on the relationships among classes \cite{maggini2012learning,prob}, on the interactions among different tasks \cite{multiview}, and on labeled regions of the input space \cite{box}, can be easily converted into constraints and embedded in the LfC learning problem (including point-wise constrains $f(x)-y=0$ on supervised pairs $(x,y)$). %
The strength of LfC is more evident when using semi-supervised data, thus enforcing constraints also on unsupervised samples.
Depending on the type of knowledge, constraints can be convex or non-convex, enforced in a soft or hard way \cite{gnecco2015learning}. 

To the best of our knowledge, LfC has always been conceived as a \textit{centralized} framework, where constraints (i.e., knowledge), data, and the learned predictors are all handled within the same computational unit. This paper studies the extension of LfC to the \textit{distributed} setting, where multiple computational nodes, connected over the network, contribute to the learning process. This setting is inspired by the nowadays organization of data and knowledge, where it is extremely common to participate to communities over the net, sharing some resources (e.g., public photos on social networks), keeping other local (e.g., private pictures taken with a personal smartphone, saved on the cloud), and having the need of developing (and eventually sharing) customized or more robust services that might benefit both from private data and public data taken from the net (e.g., a recognizer of pictures of a custom type).
Our goal consists in formulating a distributed implementation of LfC with a generic structure that covers the described setting and that could be further extended emphasizing more specific aspects, such as the ones related to privacy-preserving methods~\cite{wainwright2012privacy,chaudhuri2011differentially,rajkumar2012differentially,mtwo}.
The generality of LfC prevents the direct application of many distributed optimization approaches (see \cite{yin2018gradient} and references therein), since we need to support hard, soft, convex and nonconvex constraints.
There has been several recent progresses in the field of distributed constrained optimization \cite{wai2016projection,di2016next,tatarenko2017non,margellos2017distributed}, and the Asynchronous Method of Multipliers (ASYMM) \cite{farina2018distributed,farina2018asynchronous} offers the capability of dealing with convex and nonconvex constraints that are locally defined in computational nodes. Moreover, ASYMM has been proved to be equivalent to a centralized instance of the Method of Multipliers \cite{bertsekas2014constrained}, thus inheriting 
the properties of its centralized counterpart. 

It is worth observing that recently there has been an increased interest in
distributed learning scenarios (see, e.g.,~\cite{dean2012large,kraska2013mlbase,li2014scaling,xing2015petuum}). Specific frameworks have been studied, like the one of federated learning~\cite{konevcny2016federated,smith2017federated}, and several algorithms have been proposed \cite{boyd2011distributed,duchi2012dual,mfour}.  However, distributed learning is usually intended in the sense of \emph{learning from distributed datasets}, and central servers are required to perform at least a part of the learning process. Conversely, in this paper we consider a scenario in which not only data, but also knowledge is distributed in the
network. Moreover, we exploit a fully distributed architecture, in which no central computational unit is required.  

The main contribution of this work is to tailor the ASYMM algorithm to the aforementioned LfC distributed setting, showing how constraints can be used as a bridge between shared and private resources. As a proof-of-concept, the model is applied to two real-world problems: digit image classification and document classification. In both cases, we consider semi-supervised data, constraints on supervised examples, and constraints devised from FOL formulas. The results show that, in this distributed setting, FOL-based constraints improve the quality of the private classifiers, and local and shared constraints are asymptotically fulfilled.

\section{Learning from constraints}
\label{sec:lfc}

In the framework of LfC we consider the problem of finding the most suitable vector function $f:=\left[f_1,\ldots,f_{F}\right]\in \mathcal{F}$  subject to a set of constraints that models the available knowledge on the considered problem. $\mathcal{F}$ is a space of functions from $\mathcal{X}\subset\mathbb{R}^{d}$ (being $d$ the dimensionality of the input data) to $\mathbb{R}^{F}$ where a regularity measure is defined, and each $f_i$ is referred to as ``task function'' (for example, a classifier of a certain class).
It is pretty common to enforce \textit{point-wise} constraints, i.e., constraints applied to $f$ evaluated on a given collection of data points $\mathcal{D}$, and to consider both the \textit{bilateral} and/or \textit{unilateral} cases, that we denote by
\begin{equation}
	\Phi\left(f \mid \mathcal{D}\right)=0,\quad \check{\Phi}\left(f \mid \mathcal{D}\right)\leq 0,
	\label{a}
\end{equation}
respectively. Notice that $\Phi$ and $\check{\Phi}$ compactly indicate vectors of constraints\footnote{For simplicity, all the constraints are applied to the same $\mathcal{D}$, but our approach also holds when different constraints operate on different data. We will sometimes replace the vector $f$ with an explicit list of functions.}.
We consider $\mathcal{D}$ to be partitioned into a collection of points for which a label $y \in Y$ is known and a set of unlabeled points, respectively collected in $\mathcal{\hat{D}}$ and $\mathcal{\tilde{D}}$,
\begin{equation}
	\mathcal{D} = \mathcal{\hat{D}} \cup \mathcal{\tilde{D}} \ .
	\label{data}
\end{equation}
A popular category of constraints that is frequently exploited in LfC is given by polynomials derived from First-Order Logic formulas \cite{gori2013constraint} . In particular, each task function $f_i$ is assumed to implement the activation in $[0,1]$ of a predicate that describes a property of the considered environment, and FOL formulas represent relationships among such properties, i.e., among the tasks in $f$. FOL formulas are then converted into numerical constraints using Triangular Norms (T-Norms, \cite{klement2013triangular}), special binary functions that generalize the conjunction operator $f_i \land f_j$. For example, we might know that, in the considered environment, $f_1(x) \land f_2(x) \Rightarrow f_3(x)$, $\forall x\in\mathcal{D}$. This information is converted into a bilateral constraint of Eq. (\ref{a}), that in the case of the product T-Norm is $f_1(x)f_2(x)(1-f_3(x))=0$, and applied to all the data points of $\mathcal{D}$ (see \cite{gori2013constraint} for more examples).
In this paper, we assume $f$ to be a generic neural network. As regularity measure we use the squared norm of the weights, leading to the popular weight decay term (for simplicity, we avoid reporting this term in the following equations).

Depending on the nature of the constraints, it could be necessary to enforce some of them in a \textit{hard} way, and others in a \textit{soft} manner \cite{gnecco2015learning}. While supervisions on examples are generally subject to noise (suggesting a penalty-based soft enforcement), there might be structural or environment-related conditions that must be enforced in a \textit{hard} way. 
In the rest of the paper, we will use the notation $\Phi,\check{\Phi}$ to refer to those constraints that must be enforced in a hard way, while $\psi$ indicates the sum of the penalty functions associated to the soft constraints\footnote{The definitions of $\psi$ can include weighting terms to give different importance to the different soft constraints.}.
Moreover, the choices of both the form of the constraints of \eqref{a} and of the form of $f$ usually end up in generating constraints that are \textit{nonconvex} with respect to the model parameters that are subject of optimization. This consideration holds even more strongly when we select $f$ to be a generic neural net.

\section{Distributed framework}
\label{sec:distributed}
\begin{figure}
\centering
\includegraphics[width=0.4\textwidth]{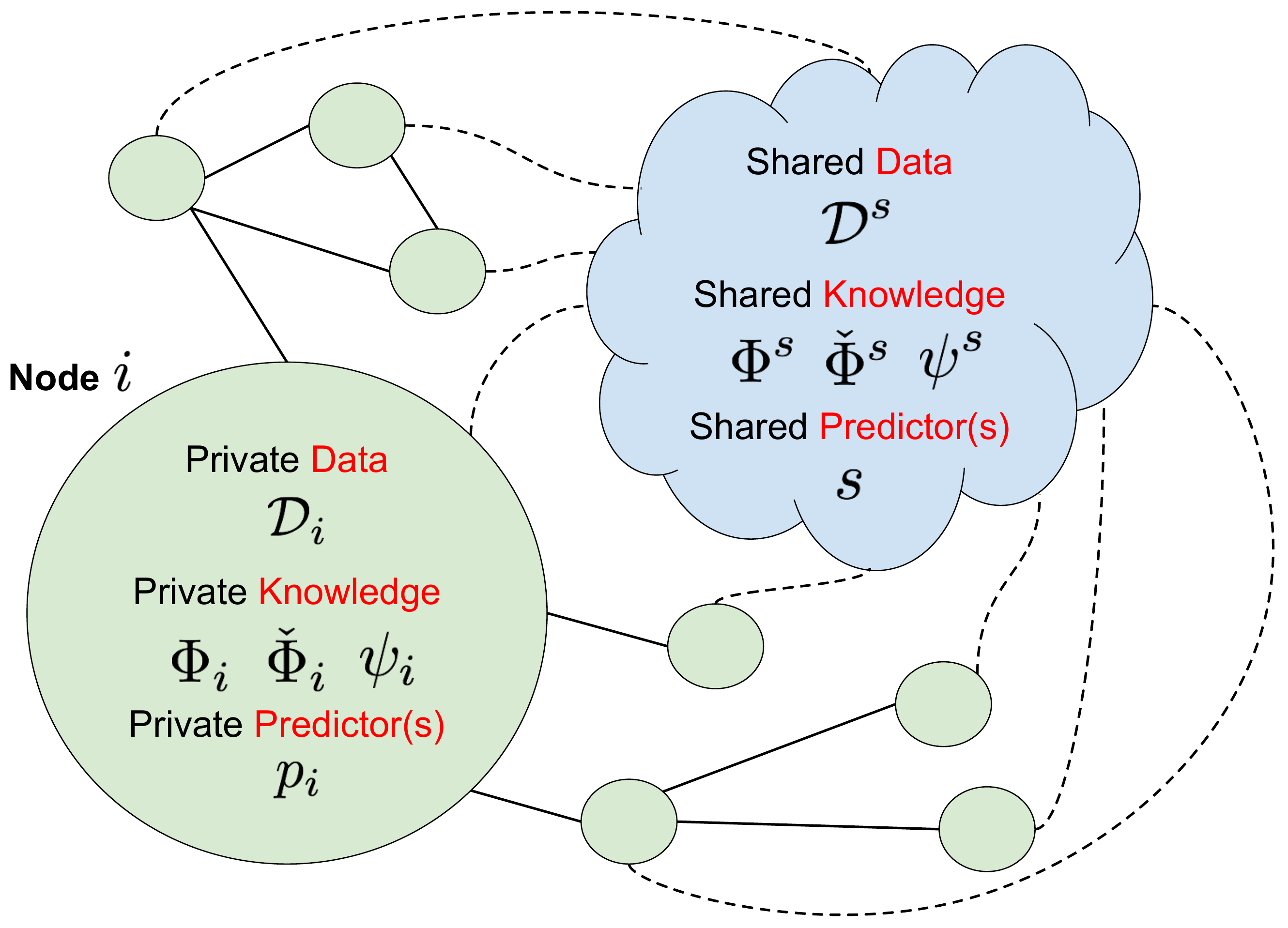}
\vskip -2mm
\caption{An illustrative view of the distributed framework and of the notation used in the paper. The graph $\GG=(\VV,\EE)$ is defined by the greenish nodes (belonging to $\VV$) and the solid connections between them (belonging to $\EE$). The blue cloud only includes shared elements, and it is accessible by all the nodes of the graph.}
\label{fig:dist}
\end{figure}

When moving to the distributed setting, we consider $N$ computational nodes connected over a network, whose underlying connectivity structure can be represented through an undirected and connected graph $\GG=(\VV,\EE)$, where $\VV$ is the set of nodes and $\EE$ is the set of node-to-node connections. Nodes might have some specific requirements in terms of the resources they want to keep \textit{private} (local) and the ones they want to \textit{share} with the other nodes. We distinguish among three types of resources: \textit{data} (i.e, the available data points), \textit{knowledge} (i.e. constraints), and \textit{predictors} (the outcome of the learning process, i.e., the task functions in $f$). Figure \ref{fig:dist} illustrates the distributed framework with privacy conditions, where we can distinguish the node-private resources and the shared ones, accessible by all the nodes. From the notation point of view, the subscript $i$ indicates a private resource of the $i$-th node, while the superscript $s$ is used to refer to shared resources.

More formally, for each node $i$ we consider some private data $\mathcal{D}_i$, private knowledge $\Phi_i,\check{\Phi}_i,\psi_i$, and some private predictors modeled with a vector function $p_i(x,w_i)$, where $w_i$ are the learnable parameters of the neural network. Similarly, for the whole network we define shared data $\mathcal{D}^s$, shared knowledge $\Phi^s,\check{\Phi}^s,\psi^s$, and some shared predictors implemented by the vector function $s(x,w^s)$, with parameters $w^s$. Then, the merged collection of all the data (Eq. (\ref{data})) and the set of all the model parameters (shared and private) are 
\begin{align}
\DD=\left(\bigcup_{i=1}^N \DD_i \right) \cup \DD^s, & & w=\left(\bigcup_{i=1}^N w_i \right) \cup w^s\ .
\end{align} 
The centralized optimization problem we have to solve is
\begin{equation}\label{pb:problem}
	\begin{aligned}
		& \m_{w}
		& & \sum_{i=1}^{N}\psi_i\left(p_i,s\mid\DD_i \cup \DD^s\right) + \psi^s\left(s\mid\DD\right)\\
		& \st
		& & \Phi_i\left(p_i,s\mid\DD_i \cup \DD^s \right)=0, &\hskip -1.5cm \forall i=1,\ldots,N\\
		& & & \check{\Phi}_i\left(p_i,s\mid\DD_i \cup \DD^s \right)\leq 0, &\hskip -1.5cm \forall i=1,\ldots,N\\
		& & & \Phi^s\left(s\mid \DD \right)=0, &\\
		& & & \check{\Phi}^s\left(s\mid\DD \right)\leq 0 &		
	\end{aligned}
\end{equation}
where constraints involve the just introduced vector functions $p_i \in \mathcal{F}$, $i=1,\ldots,N$, and $s\in \mathcal{F}$.
Notice that the private constraints can involve both private and shared predictors, thus bridging shared and local resources. Due to their shared nature, constraints $\Phi^s,\check{\Phi}^s,\psi^s$ can be enforced in all the available data.

Let us define $\wis$ as a local copy of $w^s$ made by node $i$ and $w{\mid}_i=w_i\cup\wis$.
Following the same intuition behind \cite{farina2018distributed}, we rewrite problem~\eqref{pb:problem} in an equivalent form, exploiting the connectedness  of $\GG$,
\begin{equation}\label{pb:distributed_problem}
	\begin{aligned}
		& \m_{w{\mid}_1,\ldots,w{\mid}_N}
		& & \sum_{i=1}^{N}\left(\psi_i\left(p_i,\si\mid\DD_i \cup \DD^s\right)  \vphantom{\frac{1}{N}} \right.& \hskip -0.45cm \left.+ \frac{1}{N}\psi^s\left(\si\mid \DD^s\right) \vphantom{\frac{1}{N}} + \psi^s\left(\si \mid \DD_i \right) \right)\\		
		& \st
		& & \wis = \wjs, &\forall(i,j)\in\EE\\
		& & & \Phi_i\left(p_i,\si\mid\DD_i\cup \DD^s\right)=0, &\forall i\in\VV\\
		& & & \check{\Phi}_i\left(p_i,\si\mid\DD_i\cup \DD^s\right)\leq 0, &\forall i\in\VV\\
		& & & \Phi^s\left(\si \mid \DD_i\cup \DD^s\right)=0, &\forall i\in\VV\\
		& & & \check{\Phi}^s\left(\si\mid\DD_i\cup \DD^s\right)\leq 0 &\forall i\in\VV	
	\end{aligned}
\end{equation}
where $\si=s(x,\wis)$. The first constraint ensures consistency of the local copies over $\GG$, and the last two constraints (involving shared resources) are now replicated $N$ times, splitting the private portion of the data. The objective function has been regrouped in order to be a summation over the index $i$, paying attention to differently weigh $\psi^s$ when applied to shared or private data. This formulation of the problem can be more easily partitioned among the nodes and it helps in the application of a distributed optimization algorithms.

\section{ASYMM algorithm}
\label{sec:ASYMM}
The Asynchronous Method of Multipliers (ASYMM) is a distributed optimization algorithm that has no central authorities, and that solves constrained optimization problems in which both local cost functions and constraints can be nonconvex. Thus, it is a well suited method for solving problem~\eqref{pb:problem} in a distributed way, when rewritten in the form~\eqref{pb:distributed_problem}.

The idea behind ASYMM is rooted around the concept of computational units that wake up asynchronously at different time instants, perform some operations, and broadcast their local copies of \textit{shared} parameters $\wis$ (and some other variables) to their neighbors $\NN_i$. We assume that each node keeps waking up indefinitely and the time interval between two consecutive awakenings is bounded for all nodes. Moreover, we assume for simplicity that it cannot happen for two nodes to be awake in the same time instant. When nodes $i$ wakes up, it performs a gradient descent step on a \textit{locally defined} augmented Lagrangian until every neighboring node matches a convergence criterion based on a node-defined tolerance $\epsilon_i$. By doing so, the nodes collectively approach a stationary point of the entire augmented Lagrangian of the considered optimization problem. 
The convergence check on the augmented Lagrangian is performed by the nodes in a distributed way, using a logic-AND algorithm (see~\cite{farina2018distributed}). When a node gets aware of the convergence condition, it performs one ascent step on its local multiplier vector and it increases its penalty parameters. After a node has received
the updated multipliers and penalty parameters associated to \textit{shared} constraints from all its neighbors on $\GG$, it starts over a new Lagrangian minimization. Under suitable technical assumptions, it can be shown that the computational units collectively converge to a local minimum of problem (\ref{pb:problem}) which satisfies all the constraints. Moreover \textit{private} resources are never passed over the net.

In order to devise a specialized version of ASYMM for problem~\eqref{pb:distributed_problem}, we need to introduce the corresponding augmented Lagrangian.
Let $\nu_{ij}$ and $\rhoij$ be the multiplier vector and penalty parameter
associated to the equality constraint $\wis = \wjs$. We compactly define
$\nu_i=[\nu_{ij}]_{j\in\NN_i}$, $\rho_i=[\rhoij]_{j\in\NN_i}$. Similarly, let $\lambda_{i}$, $\lambda_i^s$
and $\peqi$, $\peqi^s$ (resp. $\mu_{i}$, $\mu_i^s$ and $\pini$, $\pini^s$) be the multiplier and penalty
parameter associated to the equality (resp. inequality)
constraint of node $i$, where the superscript $s$ denotes the association with the local copy of the shared constraints.
Moreover, let $\w=[w{\mid}_1;...;w{\mid}_N]$, and denote by $\p=[\rho_i,\peqi,\pini]_{i\in\VV}$ the vector
stacking all the penalty parameters; $\nu=[\nu_i]_{i\in\VV}$,
$\lambda=[\lambda_i, \lambda_i^s]_{i\in\VV}$ and $\mu=[\mu_i,\mu_i^s]_{i\in\VV}$
be the vectors stacking the corresponding multipliers, and, consistently, let
$\LL=[\nu;\lambda;\mu]$. 
Finally, in order to present the next equations in a simpler way, let us define two parametric functions with a compact notation:
$
	q_{c}(a,b)=\frac{1}{2c}\left(\max\{0, a+cb\}^2 -a^2\right)
$
and
$
	v_{c}(a,b)=a^\top b+\frac{c}{2}\|b\|^2,
$
where the $\max$ operator is to be intended component-wise.
Then, the augmented Lagrangian
associated to~\eqref{pb:distributed_problem} is
\begin{align}
\Lag_{\p}(\w,\LL)=&\sum_{i=1}^N \bigg\lbrace \psi_i(p_i,\si\mid\DD_i\cup \DD^s) +\frac{1}{N}\psi^s(\si\mid\DD^s) \nonumber\\
&+\psi^s(\si\mid\DD_i)+\sum_{j\in \NN_i} v_{\rhoij}(\nu_{ij},\wis-\wjs)+\nonumber\\
&+v_{\peqi}(\lambda_i,\Phi_i\left(p_i,\si\mid\DD_i\cup \DD^s\right))+\nonumber\\
&+v_{\peqi^s}(\lambda_i^s,\Phi^s\left(\si\mid\DD_i\cup \DD^s\right))+\nonumber\\
&+\mathbb{1}^\top q_{\pini}(\mu_i, \check{\Phi}_i\left(p_i,\si\mid\DD_i\cup \DD^s\right))+\nonumber\\
&+\mathbb{1}^\top q_{\pini^s}(\mu_i^s, \check{\Phi}^s\left(\si\mid\DD_i\cup \DD^s\right))
\bigg\rbrace\label{eq:L} \ ,
\end{align}
where $\mathbb{1}$ is a (column) vector of ones.
 
In order to collectively minimize~\eqref{eq:L}, nodes in ASYMM need to compute a \textit{local} augmented Lagrangian. The local augmented Lagrangian for node $i$ groups all the terms in~\eqref{eq:L} depending on ${w}{\mid}_i$ and it is defined as
\begin{align}
\tLag_{\p_{\NN_i}}(w_i,w^s{\mid}_{\NN_i},\LL_{\NN_i})
=& \psi_i(p_i,\si\mid\DD_i\cup \DD^s) +\frac{1}{N}\psi^s(\si\mid\DD^s)+ \nonumber\\
&+\psi^s(\si\mid\DD_i)+\sum_{j\in \NN_i} v_{\rhoij}(\nu_{ij},\wis-\wjs)+\nonumber\\
&+v_{\peqi}(\lambda_i,\Phi_i\left(p_i,\si\mid\DD_i\cup \DD^s\right))+\nonumber\\
&+v_{\peqi^s}(\lambda_i^s,\Phi^s\left(\si\mid \DD_i \cup \DD^s\right))+\nonumber\\
&+\mathbb{1}^\top q_{\pini}(\mu_i, \check{\Phi}_i\left(p_i,\si\mid\DD_i\cup \DD^s\right))+\nonumber\\
&+\mathbb{1}^\top q_{\pini^s}(\mu_i^s, \check{\Phi}^s\left(\si\mid\DD_i\cup \DD^s\right)))\label{eq:L_local}
\end{align}
where $w^s{\mid}_{\NN_i} = [\wjs]_{j\in \NN_i\cup\{i\}}$,
$\LL_{\NN_i}=[\lambda_i,\lambda_i^s,\mu_i,\mu_i^s,\nu_{i},[\nu_{ji}]_{j\in \NN_i}]$,
and $\p_{\NN_i}=[\peqi,\peqi^s,\pini,\pini^s,\rho_i, [\rho_{ji}]_{j\in\NN_i}]$.

Finally, we define a local binary matrix $S_i\in\{0,1\}^{d_G\times d_i}$ for each node, where $d_G$ is the graph diameter and $d_i=|\NN_i|+1$. Such a matrix is used to perform the distributed logic-AND algorithm, which is a building block of ASYMM (see~\cite{farina2018distributed}).

Given the above definitions, the ASYMM algorithm applied to problem~\eqref{pb:distributed_problem} is reported in Algorithm~\ref{alg:ASYMM}, being $\alpha_i>0$ the stepsize selected by node $i$.

\begin{algorithm}
	\caption{ASYMM}\label{alg:ASYMM}
	\begin{algorithmic}
		\init $w{\mid}_i$, $\LL_i$, $\NN_i$, $\p_i$, $S_i=\mathbf{0}_{d_G\times d_i}$, $M_{done}$ = 0.
		\item[]

		\vskip -2mm\awake
		\If{$\prod_{b=1}^{d_i}S_i[d_G,b]\neq 1$ \textbf{and} \textbf{not} $M_{done}$}\\
		\vspace{-1ex}
		\State $\wis\gets \wis-\alpha_i\nabla_{\wis}\tLag_{\p_{\NN_i}}(w_i,w^s{\mid}_{\NN_i},\LL_{\NN_i})$\\
		\vspace{-1ex}
		\If{$\|\nabla_{\wis}\tLag_{\p_{\NN_i}}(w_i,w^s{\mid}_{\NN_i},\LL_{\NN_i})\|{\leq}\epsilon_i$} $S_i[1,d_i]\gets 1$\\
		\vspace{-1.5ex}
		\EndIf
		\State $S_i[l,d_i]\gets\prod_{b=1}^{d_i}S_i[l-1,b]$ for $l=2,...,d_G$\\
		\vspace{-1.5ex}
		\State \broad $\wis$, $S_i[:,d_i]$ to all $j\in \NN_i$
		\EndIf
		\item[]
		
		\vskip -3mm\If{$\prod_{b=1}^{d_i}S_i[d_G,b]=1$ \textbf{and} \textbf{not} $M_{done}$}\\
		\vspace{-1ex}
		\State $\nu_{ij}\gets\nu_{ij}+\rho_{ij}(\wis-\wjs)$ for $j\in \NN_i$\\
		\vspace{-2ex}
		\State $\lambda_{i}\gets \lambda_{i}+\peqi \Phi_i\left(p_i,\si\mid\DD_i\cup\DD^s\right)$\\
		\vspace{-2ex}
		\State $\lambda_{i}^s\gets \lambda_{i}^s+\peqi^s \Phi^s\left(\si\mid\DD^s\right)$\\
		\vspace{-2ex}
		\State $\mu_{i}\gets \max\{0,\, \mu_{i}+\pini \check{\Phi}_i\left(p_i,\si\mid\DD_i\cup\DD^s\right)\}$\\
		\vspace{-2ex}
		\State $\mu_{i}^s\gets \max\{0,\, \mu_{i}^s+\pini^s \check{\Phi}^s\left(\si\mid\DD^s\right)\}$\\
		\vspace{-1ex}
		\State update $\peqi$, $\pini$ and $\rho_i$
		\State $M_{done}$ $\gets$ 1
		\State \broad $\nu_{ij}$, $\rhoij$ to $j \in \NN_i$
		\EndIf
		\item[]

		\vskip -3mm\idle
		\State \textbf{If }$S_j[:,d_j]$ received from $j\in \NN_i$ and not already received some new $\nu_{ji}$ \textbf{then} $S_i[l,j_i]\gets S_j[l,d_j]$ for $l=1,...,d_G$
		\State \textbf{if} $\nu_{ji}$ and $\rho_{ji}$ received from $j\in \NN_i$  set $S_i\left[d_G,:\right]\gets 1$
		\State \textbf{if} $\wjs^{new}$ received from $j\in \NN_i$, update $\wjs\gets \wjs^{new}$
		\If{$M_{done}$ \textbf{and} $\nu_{ji}$ received from all $j\in \NN_i$}
		\State $M_{done}$ $\gets$ 0, $S_i\gets\mathbf{0}_{d_G\times d_i}$, update $\epsilon_i$
		\EndIf
	\end{algorithmic}
\end{algorithm}

In~\cite{farina2018distributed}, it has been shown that the distributed Algorithm~\ref{alg:ASYMM} is equivalent to a centralized version of the Method of Multipliers in which the primal update is carried out by means of a \emph{block-coordinate gradient descent} on the augmented Lagrangian (e.g., see \cite{wright2015coordinate} for a survey on coordinate descent algorithms). Specifically, there exists a sequence of (centralized) block-coordinate gradient descent steps that returns the same sequence of estimates $\wis$ as those computed by ASYMM. This equivalence property implies that ASYMM inherits all the convergence properties of the centralized block-coordinate Method of Multipliers. In particular, under technical assumptions on the local augmented Lagrangian, similar to those adopted in the centralized case (see~\cite{bertsekas2014constrained}), the estimates $\wis$ generated by ASYMM converge to a local minimum. A key point to establish this result is to bound the norm of the gradient of the augmented Lagrangian~\eqref{pb:distributed_problem} by a function of the local tolerances~$\epsilon_i$ employed in Algorithm~\ref{alg:ASYMM}. The interested reader is referred to~\cite{farina2018distributed} for a thorough theoretical analysis of ASYMM.

\section{Experiments}
\label{sec:validation}
We evaluate the numerical application of our approach to two different distributed environments, focussing on digit recognition and document classification, respectively.

\subsection{Digit Recognition} We consider a network composed by $10$ nodes, indexed from $0$ to $9$. In the context of digit recognition, each node aims at learning to recognize a precise digit given its image $x$, where we assume that node $i$ learns to recognize digit $i$. Notice that each node could also learn to recognize more than one digit. We consider only one digit per node for the sake of presentation. The $i$-th recognizer is a \textit{private function} $p_i(x,w_i) \in [0,1]$, and the $i$-th node has the use of \textit{private data} $\DD_i=\hat{\DD}_i\cup\tilde{\DD}_i$ composed of positive examples of such digit and negative examples of other digits (labeled with $y=1$ and $y=0$, respectively, and collected in $\hat{\DD}_i$), and unsupervised examples (belonging to $\tilde{\DD}_i$). No shared data are considered, so that $\DD=\bigcup_{i=0}^{9} \DD_i$. 
All the nodes of the network have access to a \textit{shared function} with two scalar outputs $s(x,w^{s}) = \left[ s_{0}(x,w^{s}), s_{1}(x,w^{s})\right] \in [0,1]^2$, that predicts whether $x$ is even (first output) or odd (second output)\footnote{We used two outputs to emphasize the role of the mutual-exclusivity constraints that we will introduce shortly.}.

\begin{table}[]
\caption{Polynomial constraints from logic statements (prod. T-Norm)}
\footnotesize
\begin{center}
\begin{tabular}{c|c}
predicate & polynomial form\\
\hline 
a $\Rightarrow$ b & $a(1-b)=0$\\\hline
$\lnot$ (a $\wedge$ b) & $ab=0$\\\hline
(a $\wedge$ b) $\Rightarrow$ c & $ab(1-c)=0$\\\hline
a $\veebar$ b & $
	a+b-1=0, 
	ab=0$
\end{tabular} 
\end{center}
\label{tab:logic}
\vskip -3mm
\end{table}

Fitting the labeled examples in node $i$ is a private soft-constraint, and it depends on $p_i$ and $\hat{\DD}_i$ only, 
\begin{equation}\label{eq:obj_func_MNIST}
	\psi_i(p_i\mid\hat{\DD}_i)=\sum_{(x,y)\in\hat{\DD}_i}\left(p_i(x,w_i)-y\right)^2 \ .
\end{equation}

Due to the private nature of $\mathcal{D}_i$, each node has no information that it can directly use to learn $s$ in discriminative way. However, each node has \textit{private knowledge} about the fact that its associated digit is either even or odd, and all the nodes have access to the \textit{shared knowledge} that $s_{0}$ and $s_{1}$ are mutually exclusive. Using FOL, we get the following universally quantified formulas (for the sake of simplicity, we skip the arguments of $p_i$ and $s$),
\begin{eqnarray*}
p_i \Rightarrow s_{0} & \text{for } i=0,2,4,6,8\\
p_i \Rightarrow s_{1} & \text{for } i=1,3,5,7,9\\
s_{0} \veebar s_{1}. & 
\end{eqnarray*}

Using the product T-Norm (see Table~\ref{tab:logic}), we convert the formulas into polynomial constraints that, following the notation of problem (\ref{pb:distributed_problem}), become
\begin{align*}
 \Phi_i(p_i,\si | \DD_i)&= p_i(x,w_i)(1-s_0(x,\wis))=0,
  &{\forall x\in\DD_i,\,i=0,2,4,6,8}\\
  \Phi_i(p_i,\si | \DD_i)&= p_i(x,w_i)(1-s_1(x,\wis))=0,
  &{\forall x\in\DD_i,\,i=1,3,5,7,9}\\
 \Phi^s(\si | \DD_i)&=\left[\begin{aligned}
&s_0(x,\wis) + s_1(x,\wis)-1\\
&s_0(x,\wis)s_1(x,\wis)
\end{aligned}\right] =0, & \forall x\in\DD_i.
\label{a1}
\end{align*}

We selected the popular MNIST dataset to test our algorithm in the proposed scenario. MNIST consists of black and white images of handwritten digits of size $28\times 28$ pixels. Each image is represented through a normalized-flattened vector $x\in[0,1]^{784}$, in which each entry is a pixel intensity.
The dataset comes divided in a training set and a test set, which consist of $60000$ and $10000$ labeled samples respectively. 

In order to generate the private data $\DD_i=\hat{\DD}_i\cup\tilde{\DD}_i$ we proceeded as follows. We randomly selected $6000$ training examples, evenly distributed among classes, and we built each $\hat{\DD}_i$ from them. In particular, we selected all the $600$ images of digit $i$ as positive examples, and $600$ images of digits $\neq i$ as negative examples (evenly distributed among digits $\neq i$, and such that there is no overlap among the negative examples of the different $\hat{\DD}_i$'s). The unsupervised sets $\tilde{\DD}_i$, $i=0,\ldots,9$ were implemented by selecting the $54000$ training examples not involved in the previous operations, and evenly assigning them to each $\tilde{\DD}_i$ with no overlap (keeping the original class distribution).
We remark that this setting is different from the one that is commonly assumed in semi-supervised classification on the MNIST data, in which the same data (supervised and unsupervised) is shared by all the classifiers and in which only one class is predicted for each test example \cite{kingma2014semi,wang2016scalable,miyato2018virtual}.

\begin{figure}
\centering
\includegraphics[width=0.4\textwidth]{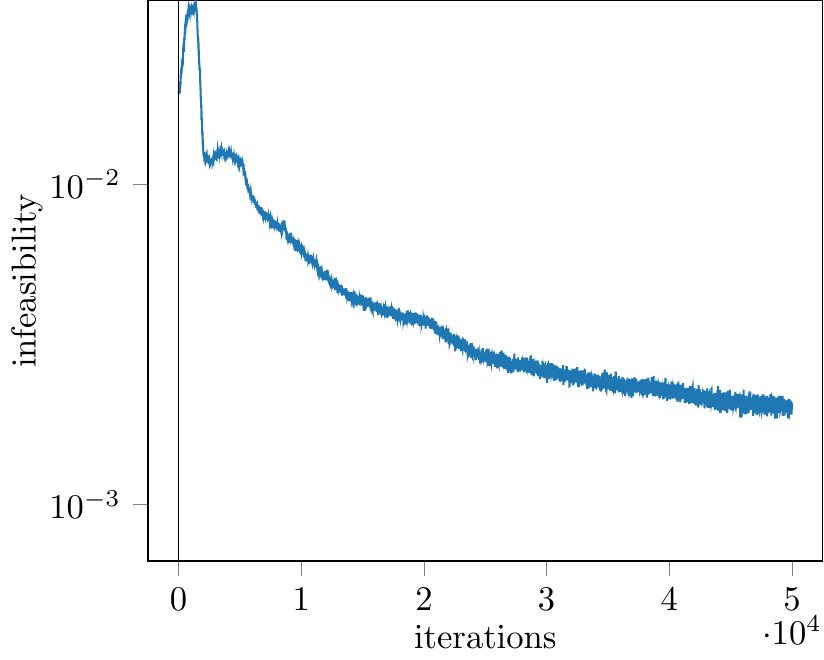} 
\includegraphics[width=0.4\textwidth]{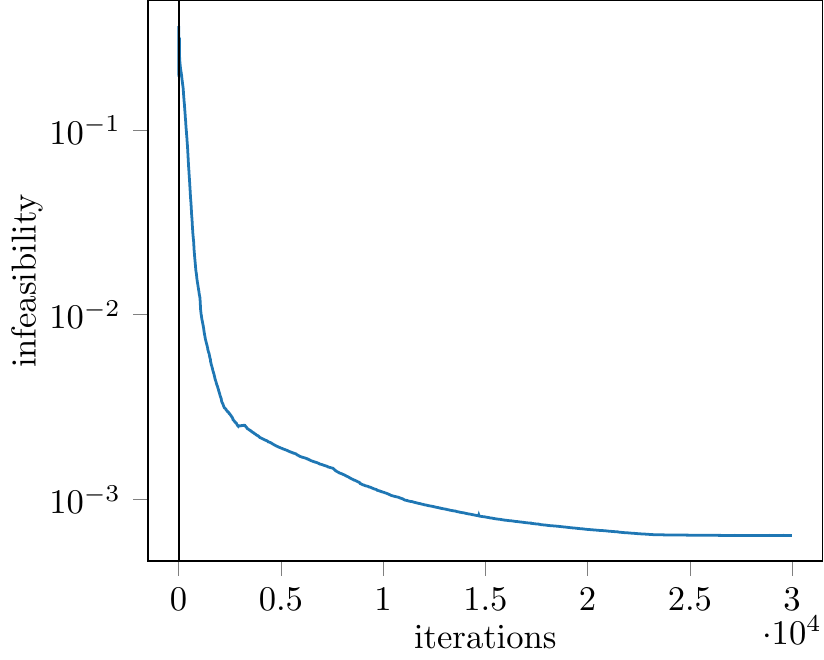} 
\caption{Average constraints violation over the evolution of the algorithm in (left) digit recognition and (right) text classification (in log scale).}
\label{fig:infeasibility_digits}
\end{figure}

We modeled each function $p_i$, $i=0,\ldots,9$, $s_0$ and $s_1$ using a simple neural architecture, that is a Multi-Layer Perceptron (MLP) with a hidden layer of $300$ units ($\tanh$ activation function) and an output unit with sigmoidal activation function. Then, we ran ASYMM by repeating $10$ times the aforementioned data generation. Each run consists of 50000 total iterations (which means 5000 awakenings per node on average). After solving the optimization problem, the learned predictors were tested using the original MNIST test set.
The mean and standard deviation of the obtained F1 scores\footnote{The F1 score is a classification performance metrics which is typically defined in terms of precision ($P$) and recall ($R$) as $F1=2\frac{PR}{P+R}$.} are reported in Table~\ref{tab:F1} (left column) for each predictor. While the results on each $p_i$ confirm that each node is reaching its goal of learning a recognizer for digit $i$, we can also see how the system is learning to correctly ($\approx0.93$) predict even and odd digits without having access to any example labeled as even or odd, but only using the hard constraints that are enforced on \textit{private data} in a \textit{distributed setting}. 
In order to experimentally verify the theoretical properties of ASYMM, we solved in a centralized way the same optimization problem considered in the presented scenario, by using the (centralized) Method of Multipliers. The results are reported in Table~\ref{tab:F1} (Centralized Semi-Supervised column) and are very close to those of the distributed implementation, the small discrepancies being due to the different orders in which block-coordinate descent steps are performed.
In the distributed scenario, in order to give an idea of the role played by unsupervised data, the simulation has been repeated without using the unsupervised data, and the results are reported in the Table~\ref{tab:F1} (right column). We can see that the F1 scores of all the classifiers are lower than in the semi-supervised scenario, confirming that the distributed implementation positively exploits the unsupervised data. Finally, in order to see how the (hard) logic constraints are asymptotically satisfied, in Figure~\ref{fig:infeasibility_digits} (left), the average constraint violation (considering all the problem constraints) is reported over the evolution of one run of ASYMM, in logarithmic scale.

\begin{table}[]
\caption{Digit classification problem, F1 score. Distributed and centralized optimization are compared. As a reference, the last column includes the results in the case in which only the supervised portion of the training data is used.}
\begin{center}
\footnotesize
\begin{tabular}{l|c|c|c|c||c|c}
	& \multicolumn{2}{c|}{Distributed} & \multicolumn{2}{c||}{{Centralized}} & \multicolumn{2}{c}{{Distributed}}\\
 	& \multicolumn{2}{c|}{Semi-Supervised} & \multicolumn{2}{c||}{{Semi-Supervised}} & \multicolumn{2}{c}{Supervised}\\\hline
	predictor & mean & std & {mean} & {std} & mean & std\\ \hline
	$p_0$ & 0.923 & 0.012 & {0.921} & {0.012} & 0.886 & 0.016 \\
	$p_1$ & 0.963 & 0.007 & {0.970} & {0.010} & 0.936 & 0.012  \\
	$p_2$ & 0.883 & 0.016 & {0.872} & {0.012} & 0.861 & 0.031  \\
	$p_3$ & 0.859 & 0.016 & {0.859} & {0.016} & 0.812 & 0.023 \\
	$p_4$ & 0.860 & 0.010 & {0.863} & {0.009} & 0.839 & 0.014 \\
	$p_5$ & 0.822 & 0.020 & {0.827} & {0.019} & 0.813 & 0.020  \\
	$p_6$ & 0.881 & 0.011 & {0.885} & {0.017} & 0.850 & 0.022 \\
	$p_7$ & 0.895 & 0.006 & {0.888} & {0.014} & 0.858 & 0.013  \\
	$p_8$ & 0.802 & 0.015 & {0.802} & {0.010} & 0.755 & 0.013 \\
	$p_9$ & 0.789 & 0.022 & {0.787} & {0.021} & 0.772 & 0.025 \\
	\hline
	$s_0$ & 0.932 & 0.008 & {0.934} & {0.010} & 0.916 & 0.016 \\
	$s_1$ & 0.932 & 0.009 & {0.934} & {0.010} & 0.915 & 0.019 
\end{tabular}
\end{center}
\label{tab:F1}
\vskip -4mm
\end{table}

\subsection{Document Classification}
In the second application, we consider the problem of document classification. 
We focus on a network of $N$ nodes, each of them associated with a category of documents ($N$ classes). Differently from the previous experiment, each node is assumed to have access to a \textit{shared predictor} $s(x,w^s)=[s_1(x,w^s),\ldots,s_N(x,w^s)]$ with $N$ outputs, that models the class-membership scores of an input document $x$, and that must be learned in a distributed setting.
What makes this task more challenging is that each node $i$ is associated to a unique document class, and it is equipped with a private dataset $\DD_i=\hat{\DD}_i\cup\tilde{\DD}_i$ of (supervised) positive-only examples from the category associated to it (i.e. all the samples in $\hat{\DD}_i$ have label $y=1$), and a set of unlabeled documents ($\tilde{\DD}_i$). 
Moreover, each node has some limited and incomplete \textit{private knowledge} on how its document category is related to the other ones.
The goal of the experiment is to make available to each node the $N$-class classifier $s$, without sharing private data, and learning from positive examples and constraints in a distributed setting.

We implemented a network of $N=6$ nodes, and picked the following $6$ document categories: \emph{clothing} ($1$), \emph{politics} ($2$), \emph{running} ($3$), \emph{shoes} ($4$), \emph{sport} ($5$) and \emph{wrestling} ($6$), where the number indicates the node index associated to each of them.
The private knowledge of each node is reported in Table~\ref{tab:text}, from which, using the polynomial forms in Table~\ref{tab:logic}, the local constraints can be easily retrieved.
As an example, following the described setup, node $4$ has the use of positive examples of category $4$ (\emph{shoes}) and some other unlabeled data. It also knows how \emph{shoes} is related with some other categories. In particular it knows that the following two relations hold: $\lnot$ (\emph{politics} $\wedge$ \emph{shoes}) and (\emph{running} $\wedge$ \emph{shoes}) $\Rightarrow$ \emph{clothing}. 
Following the rules of Table \ref{tab:logic}, the local constraint $\Phi_4$ consists of:
\begin{align*}
\Phi_4(s{\mid}_4 \mid \DD_4)
&=
\left[\begin{aligned}
&s_2(x,w^s{\mid}_4) s_4(x,w^s{\mid}_4)\\
&(s_3(x,w^s{\mid}_4) s_4(x,w^s{\mid}_4))(1-s_2(x,w^s{\mid}_4)) 
\end{aligned}\right] = 0,
& \hskip -0.75cm \forall x\in\DD_4
\end{align*}

The local objective function, instead, has the same form for all the nodes and is defined as
\begin{equation*}\label{eq:obj_func_text}
	\psi_i(s_i{\mid}\hat{\DD}_i)=\sum_{(x,y)\in\hat{\DD}_i}\left(s_i(x,w^s{\mid}_i)-y\right)^2
\end{equation*}
The considered problem is in the form of~\eqref{pb:distributed_problem} and, hence, can be solved by the ASYMM algorithm.

\begin{table}[]
	\caption{List of all the known constraints in the text classification problems. The column on the right indicates the nodes which are aware of each constraint.}
	\footnotesize
	\begin{center}
	\begin{tabular}{c|c}
	local knowledge & aware nodes\\
	\hline 

	$\lnot$ (politics $\wedge$ wrestling) & 2, 6 \\\hline
	$\lnot$ (politics $\wedge$ clothing) & 2, 1 \\\hline
	$\lnot$ (politics $\wedge$ sport) & 2, 5 \\\hline
	$\lnot$ (politics $\wedge$ running) & 2, 3 \\\hline
	$\lnot$ (politics $\wedge$ shoes) & 2, 4 \\\hline
	wrestling $\Rightarrow$ sport & 6, 5\\\hline
	(running $\wedge$ shoes) $\Rightarrow$ clothing & 3, 4, 2 \\\hline
	running $\Rightarrow$ sport & 3, 5
	\end{tabular} 
	\end{center}
	\label{tab:text}
	\vskip -3mm
\end{table}

A collection of $5180$ documents belonging to the selected categories has been obtained by crawling Wikipedia. We downloaded up to $1,000$ pages for each category, where roughly $50\%$ of the pages were taken by exploring sub-categories (limiting the depth of the exploration, and randomly deciding whether we should have considered a subcategory or not). Documents were represented by \textsc{tf-idf}, on a dictionary of $10000$ words. In this experiment, classes are not mutually exclusive. 
We marked 70\% of the documents of each class as supervised samples, while the remaining 30\% are marked as unsupervised. All the unsupervised data have been merged, randomized, and evenly assigned to the nodes (without overlap). We explored a transductive learning scenario, so the unlabeled data is also used to evaluate the quality of the learned classifiers. 

We tested two types of architectures for the classifiers $s_1,\ldots,s_6$: neural networks without hidden layers (referred to as ``single layer'') and neural networks with one hidden layer (composed by 100 units with tanh activation). Both architectures share some common properties. Namely, in their output units they have sigmoidal activation functions and a fixed negative bias ($-1$), and we enforced a strong regularization (weight decay) to better cope with the selected setting (learning from positive examples).
ASYMM has been run for $30000$ total iterations ($5000$ awakenings per node on average) and the final results are reported in Table~\ref{tab:F1_text}. 
The system is learning to classify the $6$ classes, with some low-precision results due to the lack of large discriminative information (many false positives). 
One of the classifiers with the highest scores is about the \emph{politics} class, since it is the only class which, by means of constraints, is known to be completely disjoint from the others.
The classifiers that are more involved into constraints, such as \emph{sport} and \emph{running}, yield better results than the other ones. As a matter of fact,  the other classifiers suffer from the small amount of knowledge which is injected in the system through the constraints. For example, the \emph{clothing} classifier has no information to discriminate samples from the class \emph{sport}.
Introducing a hidden layer allows the system to develop strongly non-linear decision boundaries around the given positive examples, increasing the overall performance.
As in the previous example, in order to corroborate the theoretical properties of ASYMM, we also provide the results obtained by solving the considered optimization problem in a centralized way (Table~\ref{tab:F1_text}, last column - single layer case), once again almost equivalent to those provided by ASYMM.

In order to further evaluate the proposed distributed setting, we compare the results with those of a centralized approach in which the knowledge on the relationships between the $6$ classifiers is not enforced through constraints, but by means of additional supervised data.
In particular, we merged all the training sets of the $6$ classifiers, and we enriched the supervision with additional labels that are coherent with the logic constraints of Table~\ref{tab:text}. For example, the data that are labeled as \emph{running} or \emph{wrestling} are also labeled as \emph{sport} (due to constraints 6 and 8 of Table \ref{tab:text}), while examples from the \textit{politics} class also become negative examples of the class \emph{sport} (due to constraint 3 of Table \ref{tab:text}). 
We allowed all the classifiers to have access to these data (thus violating the privacy assumption we made in the distributed setting), so that each of them can exploit an augmented collection of supervised training examples with respect to the distributed case. We remark that now classifiers have also the use of negative examples. Then, we excluded the logic constraints and the unsupervised data from the optimization (unsupervised data is not used whenever we drop the logic constraints).
Besides the two architectures considered in the set-up of Table~\ref{tab:F1_text}, we also include the case in which the classifiers are modeled as RBF networks, in order to evaluate the effect of locally supported units in this learning problem. For each classifier we considered 1000 RBF neurons and another hidden layer consisting of 100 units with tanh activation (the centers of the RBF network were estimated on an out-of-sample small set of Wikipedia documents, and shared among the nodes).
Results are reported in Table~\ref{tab:F1_text_labeled}. The RBF network performs better on some classifiers, however the overall performances are lower than those obtained with the other two architectures. 

By comparing the results reported in Table~\ref{tab:F1_text_labeled} and \ref{tab:F1_text}, it can be seen that the semi-supervised scenario (in which we exploit only positive examples, logic constraints and unsupervised data) leads, in general, to slightly better scores than those obtained in the centralized setting with artificially generated labelings. 
This comparison emphasizes the quality of the proposed distributed setting and validates the idea of sharing knowledge by means of constraints.
Finally, the average constraint violation along the evolution of ASYMM is reported in Figure~\ref{fig:infeasibility_digits} (right) for the single layer architecture,  showing that, as expected, the violation vanishes as the algorithm proceeds.

\begin{table}[]
	\caption{Precision (P), Recall (R) and F1 score on document classification, {learning from positive examples, unsupervised data, and logic constraints}.}
	\footnotesize
	\begin{center} 
		\resizebox{\columnwidth}{!}{%
		\begin{tabular}{l|c|c|c|c|c|c|c|c|c}
		& \multicolumn{3}{c|}{{Distributed}} & \multicolumn{3}{c|}{Distributed} & \multicolumn{3}{c}{{Centralized}}\\
		& \multicolumn{3}{c|}{{1 Hidden Layer}} & \multicolumn{3}{c|}{Single Layer} & \multicolumn{3}{c}{{Single Layer}}
		\\\hline
		predictor & {P} & {R} & {F1} & P & R & F1 & {P} & {R} & {F1}\\ \hline
		$s_1$, clothing & {0.366} & {0.942} & {0.527} & 0.344 & 0.893 & 0.497 & {0.345} & {0.892} & {0.498}\\
		$s_2$, politics & {0.921} & {0.894} & {0.907} & 0.917 & 0.804 & 0.857 & {0.915} & {0.804} & {0.856}\\
		$s_3$, running & {0.554} & {0.980} & {0.709} & 0.566 & 0.963 & 0.713 & {0.566} & {0.960} & {0.712}\\
		$s_4$, shoes & {0.351} & {0.792} & {0.486} & 0.304 & 0.778 & 0.437 & {0.301} & {0.777} & {0.434}\\
		$s_5$, sport & {0.793} & {0.992} & {0.940} & 0.784 & 0.991 & 0.875 & {0.782} & {0.989} & {0.873}\\
		$s_6$, wrestling & {0.477} & {0.981} & {0.641} & 0.457 & 0.970 & 0.621 & {0.455} & {0.972} & {0.620}
		\end{tabular}
		}
	\end{center}
	\label{tab:F1_text}
\end{table}

\begin{table}[]
	\caption{{Additional reference results in document classification. Predictors are independent, and they exploit a set of positive and also negative examples, obtained by artificially augmenting the original supervised portion of the training data with labels that are coherent with the logic constraints.}}
	\footnotesize
	\begin{center} 
		\resizebox{\columnwidth}{!}{%
		\begin{tabular}{l|c|c|c|c|c|c|c|c|c}
		& \multicolumn{3}{c|}{{Standard Network}} & \multicolumn{3}{c|}{{Standard Network}} & \multicolumn{3}{c}{\multirow{2}{*}{{RBF Network}}}\\
		& \multicolumn{3}{c|}{{1 Hidden Layer}} & \multicolumn{3}{c|}{{Single Layer}} \\ \hline
		predictor & {P} & {R} & {F1} & P & R & F1 & {P} & {R} & {F1} \\ \hline
		$s_1$, clothing & {0.355} & {0.940} & {0.516} & 0.328 & 0.941 & 0.487 & {0.557} & {0.949} & {0.697} \\
		$s_2$, politics & {0.923} & {0.929} & {0.921} & 0.937 & 0.907 & 0.922 & {0.882} & {0.888} & {0.882}  \\
		$s_3$, running & {0.491} & {0.990} & {0.657} & 0.490 & 0.980 & 0.653 & {0.487} & {0.996} & {0.654} \\
		$s_4$, shoes & {0.244} & {0.781} & {0.371} & 0.248 & 0.722 & 0.369 & {0.135} & {0.861} & {0.231} \\
		$s_5$, sport & {0.759} & {0.995} & {0.862} & 0.748 & 0.995 & 0.855 & {0.775} & {0.989} & {0.868} \\
		$s_6$, wrestling & {0.412} & {0.980} & {0.581} & 0.395 & 0.983 & 0.563 & {0.504} & {0.981} & {0.660} 
		\end{tabular}
		}
	\end{center}
	\label{tab:F1_text_labeled}
\end{table}

\section{Conclusions}
\label{sec:conclusions}
We proposed a distributed implementation of the framework of Learning from Constraints. We exploited the Asynchronous Method of Multipliers (ASYMM), and we implemented and evaluated a distributed setting where local (private) and shared resources (including constraints) are considered. Experiments were performed on distributed digit and document classification, confirming the quality of the proposed approach.

\end{document}